%% file: root.tex
\documentclass[letterpaper, 10 pt, conference]{ieeeconf}  

\IEEEoverridecommandlockouts                              
\usepackage{amsmath}

\usepackage{hyperref}

\usepackage{subcaption}
\usepackage{makecell}
\usepackage{array}
\usepackage{graphics} 
\usepackage{amsmath} 
\usepackage{textcomp}
\usepackage{todonotes}
\usepackage{pdfpages}
\usepackage{url}
\usepackage{mathtools}
\usepackage{bm}
\usepackage{esvect}
\usepackage{float}
\usepackage{svg}
\usepackage{booktabs}

\usepackage{amsmath}
\usepackage{siunitx}
\usepackage{placeins}
\usepackage{graphicx}
\usepackage{color,transparent}

\usepackage[utf8]{inputenc}

\usepackage[noadjust]{cite}

\let\labelindent\relax

\usepackage[inline]{enumitem}

\usepackage[T1]{fontenc}
\usepackage{multirow}

\usepackage{amsmath,amssymb}

\usepackage{todonotes}

\usepackage[normalem]{ulem}
\graphicspath{{.}}
\title{\LARGE \bf
DMMGAN: Diverse Multi Motion Prediction of 3D Human Joints using Attention-Based Generative Adversarial Network
}

\author{Payam Nikdel, Mohammad Mahdavian, Mo Chen
  \thanks{School of Computing Science, Simon Fraser University (SFU), Canada. 
          {\tt\small \{pnikdel, mmahdavi, mochen\}@sfu.ca}}
  \thanks{This work received support from Terramera Inc., the Mitacs Accelerate Program, Amii, and the CIFAR Program. M. Mahdavian received support from the SFU Graduate Deans Entrance Scholarship.}
}

\renewcommand{\baselinestretch}{0.945} 

\DeclareMathOperator*{\argmin}{arg\,min}

\usepackage{dsfont}

\begin{document}
\maketitle
\thispagestyle{empty}
\pagestyle{empty}

\begin{abstract}
Human body motion prediction is a fundamental part of many human-robot applications. Despite the recent progress in the area, most studies predict human body motion relative to a fixed joint, only limit their model to predict one possible future motion, or both. However, due to the complex nature of human motion, a single prediction cannot adequately reflect the many possible movements one can make. Also, for any robotics application, prediction of the full human body motion including the absolute 3D trajectory -- not just a 3D body pose relative to the hip joint -- is needed.  
In this paper, we try to address these two shortcomings by proposing a transformer-based generative model for forecasting multiple diverse human motions. Our model generates \textit{N} future possible body motions given the human motion history. This is achieved by first predicting the pose of the body relative to the hip joint as was done in prior work. Then, our proposed \textit{Hip Prediction Module} predicts the trajectory of the hip position relative to a global reference frame for each predicted pose frame, an aspect of human body motion neglected by previous work. To obtain a set of diverse predicted motions, we introduce a similarity loss that penalizes the pairwise sample distance. Our system not only outperforms the state-of-the-art in human motion prediction, but also is able to predict a diverse set of future human body motions including the hip trajectory.

\end{abstract}

\input{Intro}

\section{Related Work}
\input{related_work}
\input{method}

\input{experiment}

\section{conclusion and future work}
We proposed DMMGAN, a novel method to predict diverse human motions. DMMGAN combined a generative adversarial network with Transformer based encoders to generate both the trajectory and the 3D pose of human motions. 

Our implementation outperformed the previous state of the art in diverse human 3D pose prediction while also predicting the human's trajectory. 

\bibliographystyle{IEEEtran}
\bibliography{root}

\end{document}

%% file: Intro.tex
\section{Introduction}

An important ability of an intelligent system interacting with humans is to estimate plausible human body pose and trajectories in 3D space. With the advancement of artificial intelligence, there are multiple industrial applications for such algorithms in human-robot interactions (HRI)~\cite{ferrer2014bayesian}, autonomous driving~\cite{gulzar2021survey} or visual surveillance~\cite{rashid2017enhancing}. Specifically, detailed human 3D body motion prediction plays a crucial role in many robotic applications, such as robot following ahead of a human~\cite{nikdel2021lbgp,7523689} or crowd navigation~\cite{chen2020relational}. 
In the past decade, with the popularity of deep learning, sequence to sequence (seq2seq) prediction methods such as those involving Recurrent Neural Networks (RNN)~\cite{martinez2017human} have shown promising results, and have become a viable alternative to conventional human motion prediction methods~\cite{lehrmann2014efficient,sigal2012loose}

\begin{figure}[t]
\includegraphics[width=0.99\linewidth]{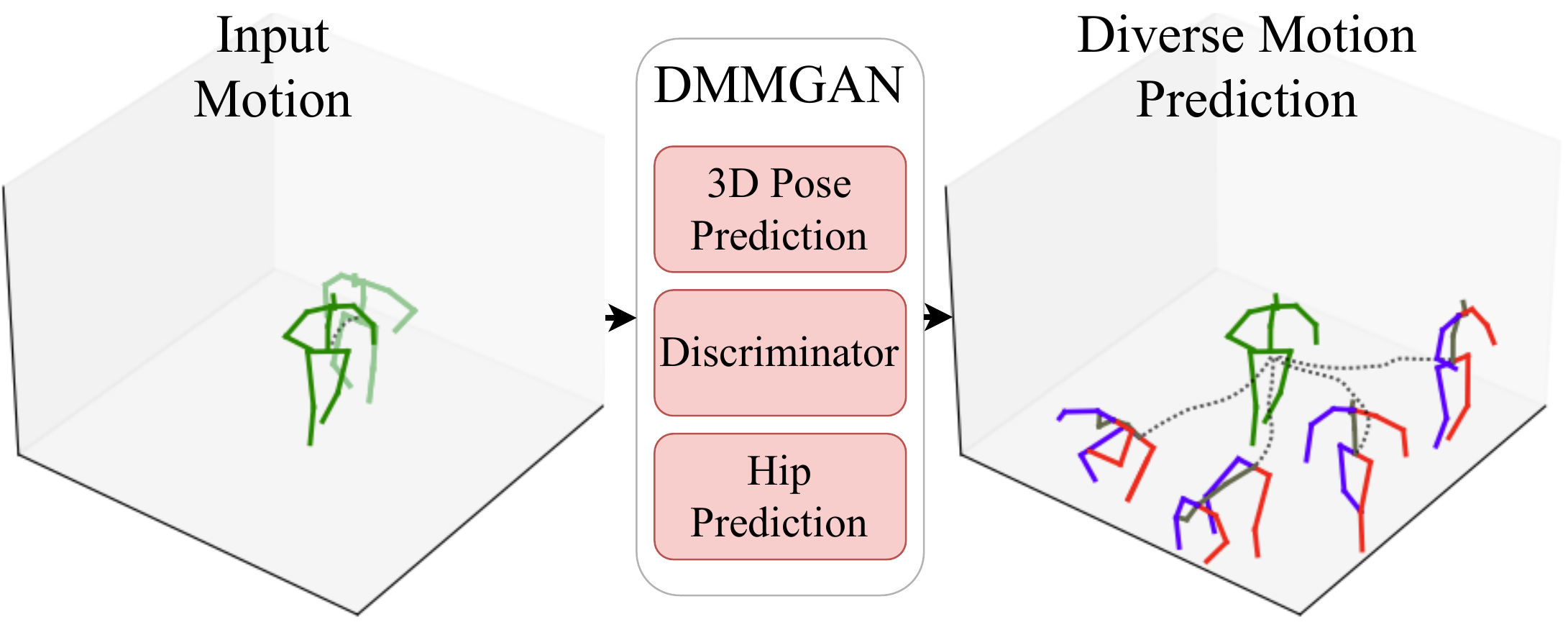}

\caption{Given a sequence of 3D human motion, our system generates a diverse set of future motions. The \textit{3D pose prediction} module generates diverse 3d poses while \textit{hip prediction} module estimates the human trajectory together forming a 3d human motion. The \textit{discriminator} module distinguishes a real 3D human motion from a generated one. }
\label{fig:cool}
\end{figure}

In general, the 3D human motion prediction problem can be divided to ~\textit{Human Pose Prediction} and~\textit{Human Trajectory Prediction}. \textit{Pose} refers to a relative position of all body joints with respect to the hip joint and \textit{Trajectory} refers to the hip joint path while the entire body moves in 3D space. For solving both problems, seq2seq models have been utilized successfully with room for improvement. Predicting a human future motion sequence can be defined as a probabilistic or deterministic problem~\cite{lyu20223d}. In probabilistic methods, similar to how our brain performs, we predict multiple future motion sequences for an observed motion sequence. Arguably, the probabilistic approach is preferred in robotic applications as it provides more assurance by considering a set of possible scenarios. However, probabilistic methods may reduce the accuracy of each individual predicted sequence. Deep generative models such as generative adversarial networks (GANs) are one of the leading deep neural network architectures that help such methods achieve reasonable accuracy. Notably, DLow~\cite{yuan2020dlow} is the state-of-the-art method that uses deep generative models and a novel sampling method for multi-future \textit{pose} predictions. 
On the other hand, deterministic methods aim to predict one single sequence more accurately, while not considering the diverse and multi-modal nature of human behaviour which reduces their practicality in some robotic applications. 
Furthermore, works on human trajectory predictions are sometimes limited due to only considering the hip movements and ignoring other joints while making a prediction, even though the joints can provide valuable information about how the hip may move in space. 

In recent years, with the emergence of transformers~\cite{attention}, many works attempted to solve the human motion prediction problem by acquiring spatio-temporal autoregressive ~\cite{aksan2021spatio} or non-autoregressive transformers~\cite{martinez2021pose}. For instance, Aksan et al.~\cite{aksan2021spatio} introduce a spatio-temporal-based transformer for 3D human motion prediction. It uses an autoregressive model that predicts human future \textit{poses}. Gonzlez et al.~\cite{martinez2021pose} improved the transformer model's inference speed by making it non-autoregressive while trading off the accuracy on long-term predictions. 
\begin{figure*}[h!]
\centering
\vspace{3mm}
\includegraphics[width=0.99\textwidth]{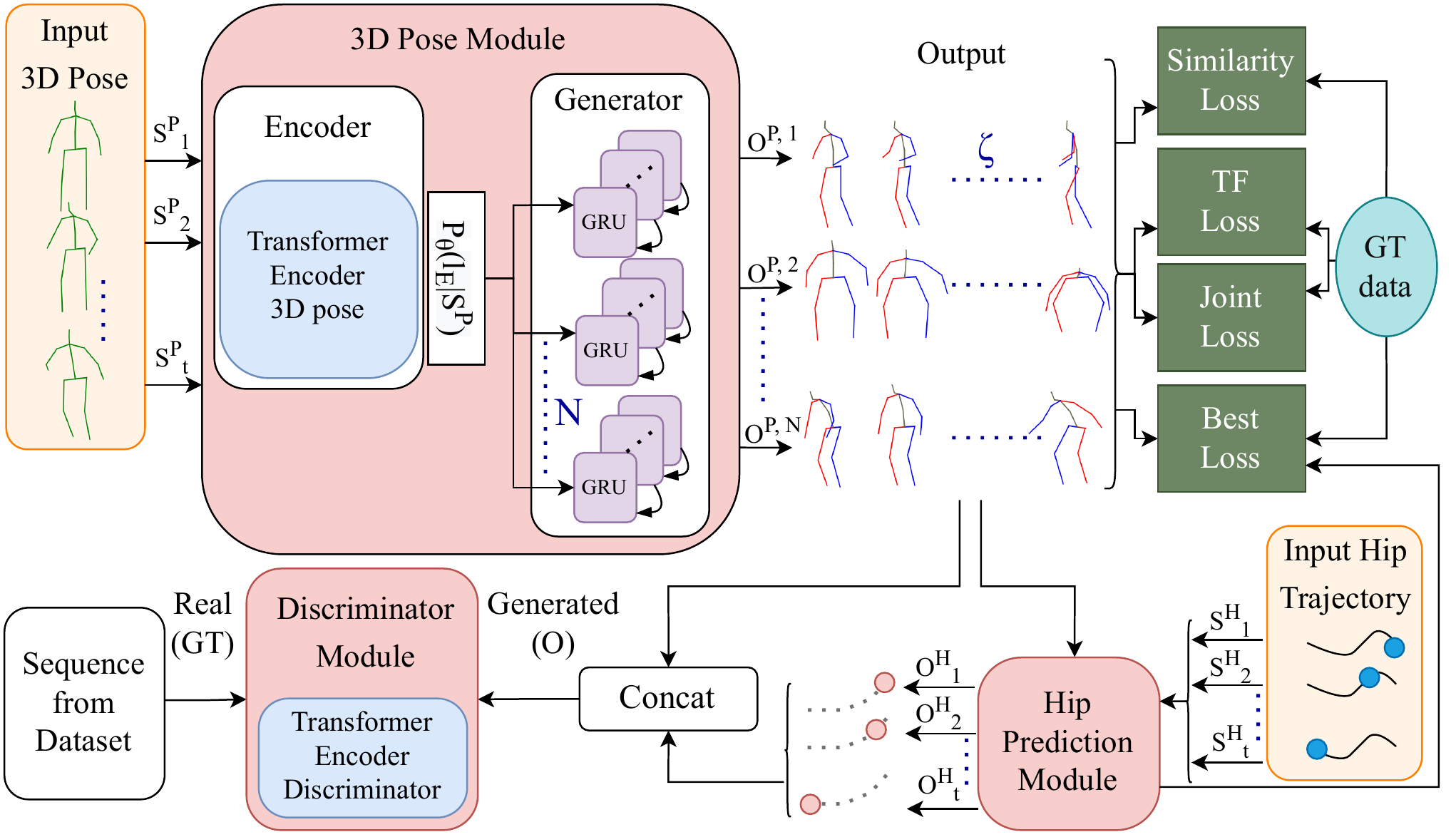}

\caption{System overview: Given a sequence of 3D human motion, our method generates $N$ future sequences of human 3D motion using a discriminator and four loss functions. Our system consists of three main parts. The first part is predicting the human 3D pose (\textit{3D Pose module}) by receiving a history of human 3D pose. The second part is the \textit{Hip Prediction} module (more details in Fig. \ref{fig:pelvis_module}) which predicts the future position of the hip joint for each of the predicted human 3D pose. Finally, the discriminator module learns the distribution of the Human 3.6M dataset by distinguishing between generated and real data. The system uses the discriminator loss to generate sequences similar to the dataset distribution while using four supervised loss functions to promote accuracy and diversity. See Fig. \ref{fig:pelvis_module} for Transformer Encoder architecture.}
\label{fig:overview}
\end{figure*}

In this work, we combine the benefit of both probabilistic and deterministic methods to provide multiple accurate predictions for both 3D human trajectory and pose. We hope this opens doors to practical use in real robotic applications.
To generate multiple future human motions, we use a conditional generative adversarial network (CGAN) with a transformer-based encoder for better encoding the observed sequence. At the end, a GRU combined with a GAN provides multiple future predictions autoregressively.


The contributions of this paper are as follows:
\begin{itemize}
\item We propose a novel deep generative architecture involving transformer-based encoders to predict a diverse set of possible human body motions.
\item We provide a real-time solution for diverse 3D human motion prediction, including both pose and trajectory prediction, which can potentially be more suitably used for robotics and autonomous car applications.
\item In addition to providing both pose and trajectory predictions, our work achieves better accuracy compared to the state-of-the-art models in standard evaluation metrics.
\end{itemize}

%% file: related_work.tex
The human pose prediction problem is divided into probabilistic and deterministic approaches. Early deterministic approaches use RNN modules for making predictions~\cite{fragkiadaki2015recurrent,martinez2017human,liu2022investigating,Zhang2021}. In recent years, Graph Convolutions Networks (GCN)~\cite{bin2020structure,li2021multiscale} and Spatio-Temporal~\cite{li2021multiscale,aksan2021spatio,medjaouri2022hr} methods attempt to improve the predictions by better learning the spatial and temporal dependencies between the joints. More recently, transformers~\cite{attention} parallelized the training process which improved the accuracy and speed of the predictions~\cite{aksan2021spatio,martinez2021pose}. 

On the other hand, probabilistic approaches gained popularity with the development of GANs. These methods~\cite{walker2017pose,lin2018human,barsoum2018hp,aliakbarian2020stochastic} usually use CGANs or Conditional Variational Autoencoders (CVAEs). As one of the state of the art methods, DLow~\cite{yuan2020dlow} generates a diverse set of samples from a pretrained deep generative model. The authors train a mapping function that samples diversely using a pretrained CVAE. To diversify the samples, they train a set of learnable mapping functions with correlated latent space that use an energy-based formulation based on pairwise sample distance. We use DLOW as one of our baselines in Section \ref{sec:experiment}. Also, Yan et al.~\cite{yan2018mt} developed a Motion Transformation Variational Autoencoder (MT-VAE) to generate multiple diverse and plausible motion sequences for facial and full body motion from an observed sequence. More recently, Agand et al.~\cite{agand2022human} developed a probabilistic and optimal approach for human navigational intent inference. All these algorithms make predictions from a human 3D pose sequence. There are a few works that perform pose prediction directly from video~\cite{zhang2019predicting}, but they can be less accurate. In this work, we assume that the human motion (pose and trajectory) are available as the input to our model, since for a real-world application we can simply get these 3D motion using hardware such as the ZED2 camera\footnote{https://www.stereolabs.com/zed-2/}.

To predict the human motion for a robotics application we need both the 3D pose and trajectory. Human trajectory prediction has been studied and implemented using RNNs~\cite{alahi2016social,song2018human} and transformers~\cite{achaji2022pretr}. But there are very few prior works that attempted to combine the human pose and trajectory predictions~\cite{stopr}. This combination can improve each individual prediction as the two parts are interdependent.


%% file: method.tex
\section{Problem Setup}

Our framework predicts a diverse set of human's motion. The input is a sequence of 3D body motion  $S = \{S_{t-\alpha}, S_{t-\alpha+1},...,S_{t}\}$ of the past human's skeleton movements capture up to the current time $t$ where $S_i\in\mathds{R}^{51}$ represents the 3D positions of 17 human joints at time $i$. 
The outputs of our system are $N$ possible sequences of future 3D human motion $O^{\gamma}_{i}=\{O^{\gamma}_{t+1},...,O^{\gamma}_{t+\zeta}\}$ where $\gamma\in {1,...,N}$ is the sequence number and $\zeta$ is the forecast duration. We divide the human 3D motion into two parts so that $S_i = (S_i^H, S_i^P)$ and $O_i = (O_i^H, O_i^P)$. The position of the hip joint is denoted by $S^H$ and $O^H$ for input and output hip trajectories. The relative positions of all joints with respect to the hip joint (called 3D pose, or just \textit{pose}), denoted by $S^P$ and $O^P$ for input and output 3D pose sequences.

\section{Method}

The overall framework of our system is summarized in Fig. \ref{fig:overview}. Our method learns to generate valid and rich human motions by leveraging the Human 3.6M dataset \cite{h36m}. It divides the prediction of human 3D motion into predicting the joints motion relative to the hip joint (3D pose) and predicting the 3D position of the hip joint in the global frame for each predicted 3D pose (human trajectory). We estimate the human trajectory by considering both the predicted 3D pose and the trajectory history. 

Specifically, we design our model to benefit from both paired and unpaired data by introducing four supervised losses and a discriminator loss respectively.
Here, given a sequence of 3D motion $\{S_{t-\alpha}, \ldots, S_t\}$, a transformer encoder learns representation of the input in a latent space.
Then, a generator uses this latent representation to output $N$ future motions. To train our system, we use 5 losses. The \textit{Best Loss} finds the best match between all the outputs and the ground truth data. The \textit{Teacher Forcing Loss} improves the final prediction by randomly feeding ground truth instead of the model prediction in the decoding phase. Similar to the Best Loss, the Teacher Forcing Loss only applies for the output that matches the most closely with the ground truth. The \textit{Similarity Loss} promotes diversity by penalizing the pairwise distance between the $N$ generated sequences, and lastly we use the \textit{Joint Loss} to encourage joint length constraints. We combine these losses with the \textit{Discriminator Loss} to generate plausible sequences matching the Human 3.6M dataset \cite{h36m}.

\subsection{Model Architecture}
Our model consists of three main modules, the first module is the \textit{3D pose module}, which generates $N$ sequences of human 3D pose (relative to the hip joint). The second module is the \textit{Hip Prediction module}, which predicts the trajectory of the hip joint in the global frame for each predicted human 3D pose. Finally the last module is the \textit{Discriminator module}, which learns the distribution of the dataset by distinguishing between the real and generated 3D sequences of human's motion. 

\subsubsection{3D Pose Module}
\begin{figure}[h]
\includegraphics[width=0.95\linewidth]{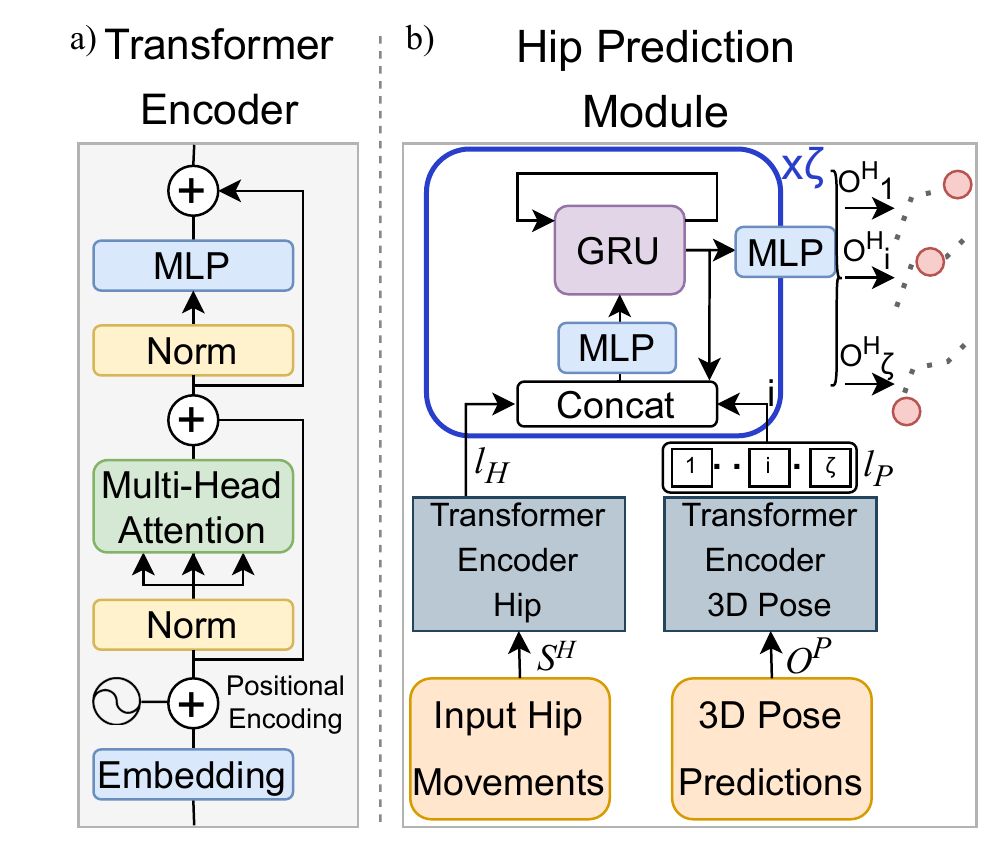}

\caption{a) The Transformer Encoder \cite{attention} and b) the \textit{Hip Prediction} module architectures. The \textit{Hip Prediction} module, estimates the hip joint positions of each predicted 3D pose by receiving the history of the hip movements and the motion predicted by the \textit{3D Pose} module.}
\label{fig:pelvis_module}
\end{figure}

The 3D Pose module consists of two parts, as shown in Fig. \ref{fig:overview}. The first part is the encoder. Given a sequence of human 3D pose $S^P$, it outputs a latent representation $l$ that encodes the past motion $P_\theta (l_E|S^P)$. Our encoder network uses a Transformer architecture, as shown in Fig. \ref{fig:pelvis_module}, to learn meaningful information over a sequence of 3D poses, similar to the model introduced by Vaswani et al.~\cite{attention}. 

The second part is the generator. It forecasts $N$ sequences of human 3D pose $O^{P, 1}, \ldots, O^{P, N}$ given the past latent representation $l_E$. Instead of using random variable as the input of the generator to forecast the future, we design it to learn a mapping from the latent representation to $N$ priors $z=f(l_E)$. Then it initializes $N$ generator networks with Gated Recurrent Units (GRU) \cite{gru}, each of which forecasts a sequence of future 3D pose based on their prior $P_{\phi n}(O^P_n|z_n), n\in \{1,...,N\}$.

\subsubsection{Hip Prediction Module}
The second module is the \textit{hip prediction} module. Given the 3D pose predictions $O^P$ and the trajectory history, it estimates the position of the hip for each predicted 3D pose. 

Fig. \ref{fig:pelvis_module} shows the architecture of the \textit{Hip Prediction} module. It uses two Transformer encoders. The first one learns a representation $l_H$ from the observed hip movements $S^H$ and the second one learns a representation $l_P$ from a predicted 3D pose sequence. If the transformer embedding has $\sigma$ dimensions and the input has a length of $\alpha$ frames and we predicted a 3D pose sequence with $\zeta$ frames, the output of $S^H$ has $\alpha \sigma$ and $S^P$ has $\zeta \sigma$ dimensions. The GRU gets the concatenation of $l_{P,i}$, $l_{H}$ and the previous output of the GRU as its input, to predict the position of the hip at time $i$ for $i=1,...,\zeta$.

\subsubsection{Discriminator Module}
The last module is the discriminator. Here we use a Transformer-based Encoder architecture shown in Fig. \ref{fig:pelvis_module}a. The input of the discriminator is the full human 3D motion, consisting of the hip trajectory and the 3D pose trajectory. The discriminator needs to distinguish between the real and the generated data (Fig. \ref{fig:overview}).

\subsection{Model Training}
During training, we exploit paired data by introducing four supervised losses to promote diversity and accuracy of the predictions. We also benefit from unpaired data by using a discriminator that learns to distinguish between the real and generated data. In the following we use the ground truth, $GT$, term to refer to the paired data, $GT^P$ and $GT^H$ to refer to the ground truth paired 3D pose and hip trajectory respectively.

\subsubsection{Discriminator Loss}
We implement the discriminator loss based on the Wasserstein Generative Adversarial Network (WGAN) \cite{wgan}. To make the training more stable we used the Gradient Penalty (GP) version of the WGAN. If $f$ is the discriminator network, the GP WGAN  critic's loss function is defined as follows:
\begin{equation}
\begin{aligned}
    &\mathcal{L}_{cWGAN}=\displaystyle \mathop{\mathbb{E}}_{O\sim P_g}[f(O)] -\displaystyle \mathop{\mathbb{E}}_{GT\sim  P_r}[f(GT)]
\end{aligned}
\label{eq:wgan}
\end{equation}
\begin{equation}
\begin{aligned}
    &\mathcal{L}_{cGP}= \mathcal{L}_{WGAN} + \lambda \displaystyle \mathop{\mathbb{E}}_{\bar{x}\sim  P_{\bar{x}}}[(||\nabla_{\bar{x}}f(\bar{x})||_2-1)^2]
\end{aligned}
\label{eq:wganpenalty}
\end{equation}

\noindent where \eqref{eq:wgan} is the original critic loss function of WGAN method and last term of \eqref{eq:wganpenalty} is the gradient penalty term. Consider a line connecting real ($P_r$) to generated ($P_g$) distributions. $P_{\bar{x}}$ is the distribution of these samples and $\lambda$ is the weight of the gradient penalty.

The second part of the discriminator loss function is the generator objective. The objective of generator is to minimise the distance between $P_g$ and $P_r$ by maximizing the expectation of the generated samples:
\begin{equation}
\begin{aligned}
    &\mathcal{L}_{g}=-\displaystyle \mathop{\mathbb{E}}_{O\sim P_g}[f(O)]
\end{aligned}
\end{equation}

\subsubsection{Best Loss}
Given a sequence of human's 3D motion, our model predicts multiple forecasts of future motions. Using the discriminator loss, these forecast would be similar to the distribution of the dataset. The Best Loss minimizes the distance between the closest prediction and the $GT$ data using mean squared error (MSE). The Best Loss is defined as follows:
\begin{align}
&   \mathcal{L}_{best}=\sum_{T=t+1}^{t+\zeta} MSE(O^\Gamma_T, GT_T) \\
\text{where } & \Gamma = \argmin_{\gamma=1,...,N}\sum_{T=t+1}^{t+\zeta} D(O^{P, \gamma}_T, GT^P_T)\\
\text{and } & D(O^\Gamma_t, GT_t)=\sum_{T=t+1}^{t+\zeta} \sum_{j=1}^{17} d(O^\Gamma_{t,j}, GT_{t,j})
\end{align}

Here, $D$ is the distance between two 3D motion predictions and $d$ is the euclidean distance between two joints.
\subsubsection{Teacher Forcing Loss}
After calculating the predicted sequence that matches with the $GT$, the \textit{Teacher Forcing} (TF) loss is calculated by randomly using the next frame from the $GT$ instead of the last prediction in the GRU (Fig. \ref{fig:overview} Generator). The TF loss can be especially useful in reducing the final displacement error as the model can learn to predict the next frames by using a combination of the $GT$ and its own predictions \cite{tf}.

\subsubsection{Similarity Loss}
We define the \textit{Similarity} loss to increase the variety of the model predictions. We first find the distance between each pair of the predicted human 3D pose. Then select the two predictions, $\Gamma_1$ and $\Gamma_2$, with the shortest distance.

\begin{equation}
\begin{aligned}
&    \Gamma_1, \Gamma_2 = \argmin_{\substack{\gamma_1\in\{1,\ldots,N\},\\\gamma_2 \in \{1,\ldots,N\} \backslash \gamma_1}}\sum_{T=t+1}^{t+\zeta} D(O^{P, \gamma_1}_T, O^{P, \gamma_2}_T)\\
\end{aligned}
\end{equation}
We can define the distance of each two joints of $\Gamma_1$ and $\Gamma_2$ by:
\begin{equation}
\begin{aligned}
&    distJoints_j = \sum_{T=t+1}^{t+\zeta} d(O^{P, \Gamma_1}_{T, j},  O^{P, \Gamma_2}_{T, j}) \\
\end{aligned}
\end{equation}
Then we apply the negative of MSE to the joints that exceed the average \textit{Similarity loss} threshold of $\epsilon$. We can define the $Similarity loss$ as follows:

\begin{align}
&    \mathcal{L}_{similarity}=-\frac{1}{16}\sum_{j=0}^{16}distPenalize_j^2, \text{ where} \\
&   distPenalize_j =
    \begin{cases}
      0 & \text{if } distJoints_j<\epsilon\\
      distJoints_j & \text{otherwise}\\
    \end{cases}
\end{align}

To make the training more stable we use the \textit{Similarity loss} only during the first \textit{M} steps of the training.
\subsubsection{Joint Loss}
As human's bone length stay the same, joint Loss works as a regularizer that helps the model by forcing it to keep the bone length similar over time.
If $V$ is the set of vertices of a graph representing all human joints and $E$ is the edges of this graph representing all human bones, then the joint loss is defined as follows:
\begin{align}
&    \mathcal{L}_{joint}=\sum_{(i,j)\in E}\sum_{\gamma=1}^{N}MSE(J^{P, \gamma}_{i,j}, J^{P, GT}_{i,j})\\
\text{where } &    J^{P, \gamma}_{i,j}=\frac{1}{\zeta} \sum_{T=t+1}^{t+\zeta}(d(O^{P, \gamma}_{T, i}, O^{P, \gamma}_{T, j})),\\
&    J^{P, GT}_{i,j}=\frac{1}{\zeta} \sum_{T=t+1}^{t+\zeta}(d(GT^{P}_{T, i}, GT^{P,}_{T, j}))
\end{align}

\subsection{Data Prepossessing}
To improve the model prediction and avoid over-fitting, we convert each 3D position in a sequence of human motion to a relative coordinate system based on the position of the hip joint at the time $t$. We also normalize each skeleton 3D pose ($\mu=0, \sigma=1$).

\subsection{Dataset}
For our experiments and training, we use the Human 3.6M dataset \cite{h36m}. Human 3.6M is a large dataset with 7 actors\footnote{There are 4 other actors without ground truth data}. For each actor, there are 15 actions that are recorded using a high-speed motion capture system at 50 Hz. Similar to DLow \cite{yuan2020dlow}, we use 17 joints skeleton and train on actors S1, S5, S6, S7 and S8 while testing on S9 and S11. For future prediction, our model observes 0.5 seconds sequence of human's body motion to forecast the next 2 seconds.

%% file: experiment.tex
\section{Experiments and Results}
\label{sec:experiment}

Our method is specifically designed to forecast 3D motions that are suitable for the autonomous car or robotics applications. It can predict the human 3D pose (position of joints relative to the hip joint) while predicting their trajectory (hip joint) separately. Most of the previous works only predict the human 3D pose without the human's hip trajectory. 

Here we designed two experiments. The first one evaluates our 3D pose prediction without the trajectory prediction module. Then in the second experiment we evaluate our full system. For both the application we used the same model (DMMGAN). Our model can run at 10 frames per seconds (FPS) on a GeForce 1080 GPU. Since most robotics applications require the observation to come with the frequency of less than 10 FPS, we train our model and the baselines using the Human3.6M \cite{h36m} at 10 FPS. For DLow and our methods we predict 10 sequence per observation ($N=10$).

To evaluate our model versus the baselines we measure the accuracy and diversity using the following metrics (we are using the evaluation metrics similar to \cite{yuan2020dlow, Yuan2}):

\subsubsection{Average Pairwise Distance (APD)}
Evaluates diversity among the predictions. We calculate the APD by averaging the pairwise distance between all pairs of 3D pose samples between the predictions. The APD is calculated as $\frac{1}{N \times (N-1)}\sum_{i=1}^{N}\sum_{j\ne i}^{N} ||O^P_i - O^P_j||$.

\subsubsection{Average Displacement Error (ADE)}
Mean squared distance between the ground-truth and the closest prediction. We define the ADE for both the 3D pose and the hip trajectory movements. We first calculate the closest prediction index, $\Gamma$, using the 3D pose predictions by: $\Gamma = \argmin_{\gamma=1,...,N}\sum_{T=t+1}^{t+\zeta} D(O^{P, \gamma}_T, GT^P_T)$. Then use this index to calculate the ADE for both the 3D pose and the trajectory: $ADE_p=\sum_{T=t+1}^{t+\zeta} D(O^{P, \Gamma_T}, GT^P_T)$ and $ADE_h=\sum_{T=t+1}^{t+\zeta} D(O^{H, \Gamma_T}, GT^H_T)$.  

\subsubsection{Final Displacement Error (FDE)} Mean squared distance between the final ground-truth and the closest final prediction. Similar to ADE, we first calculate the closest final prediction index by $\gimel= \argmin_{\gamma=1,...,N}D(O_{\gamma,{t+\zeta}}, GT_{t+\zeta})$. Then we calculate the FDE for both the 3D pose and the trajectory:  $FDE_p=D(O^{P, \gimel}_{t+\zeta}, GT^P_{t+\zeta})$ and $ADE_h=D(O^{H, \gimel}_{t+\zeta}, GT^H_{t+\zeta})$.
\subsubsection{Multi-modal ADE (MADE)} To evaluate our system's ability to generate multi-modal predictions, we used the multi-modal version of ADE  \cite{yuan2020dlow, Yuan2}. The MADE uses multi-modal $GT$ future motions by grouping similar past motions.
\subsubsection{Multi-modal FDE (MFDE)} Similar to MADE, The MFDE is the multi-modal version of FDE \cite{yuan2020dlow, Yuan2}. 
    
\begin{table}[H]
  \centering
  \begin{tabular}{cccccc}
    Approach & APD   & ADE & FDE & MADE & MFDE  \\
     &   $\uparrow$  & (m) $\downarrow$ & (m) $\downarrow$ & (m) $\downarrow$ & (m) $\downarrow$ \\
    \hline
    DMMGAN (Ours) & \bm{$5.81$} & \bm{$0.44$} &  \bm{$0.52$} &\bm{$0.54$} & \bm{$0.60$} \\
    DLow & $5.53$ & $0.48$ &  $0.61$ &$0.55$ & $0.63$\\
    STPOTR & NA & $0.50$& $0.75$ & NA & NA \\
  \end{tabular}
  
  \caption{Comparison of our systems versus two baselines for the 3D Pose experiment.}\label{tab:straight_real}
      \vspace{-5pt}
    \label{tab:exp1}
\end{table}

\begin{figure*}[ht!]
\vspace{3mm}
\centering

\includegraphics[width=0.95\linewidth]{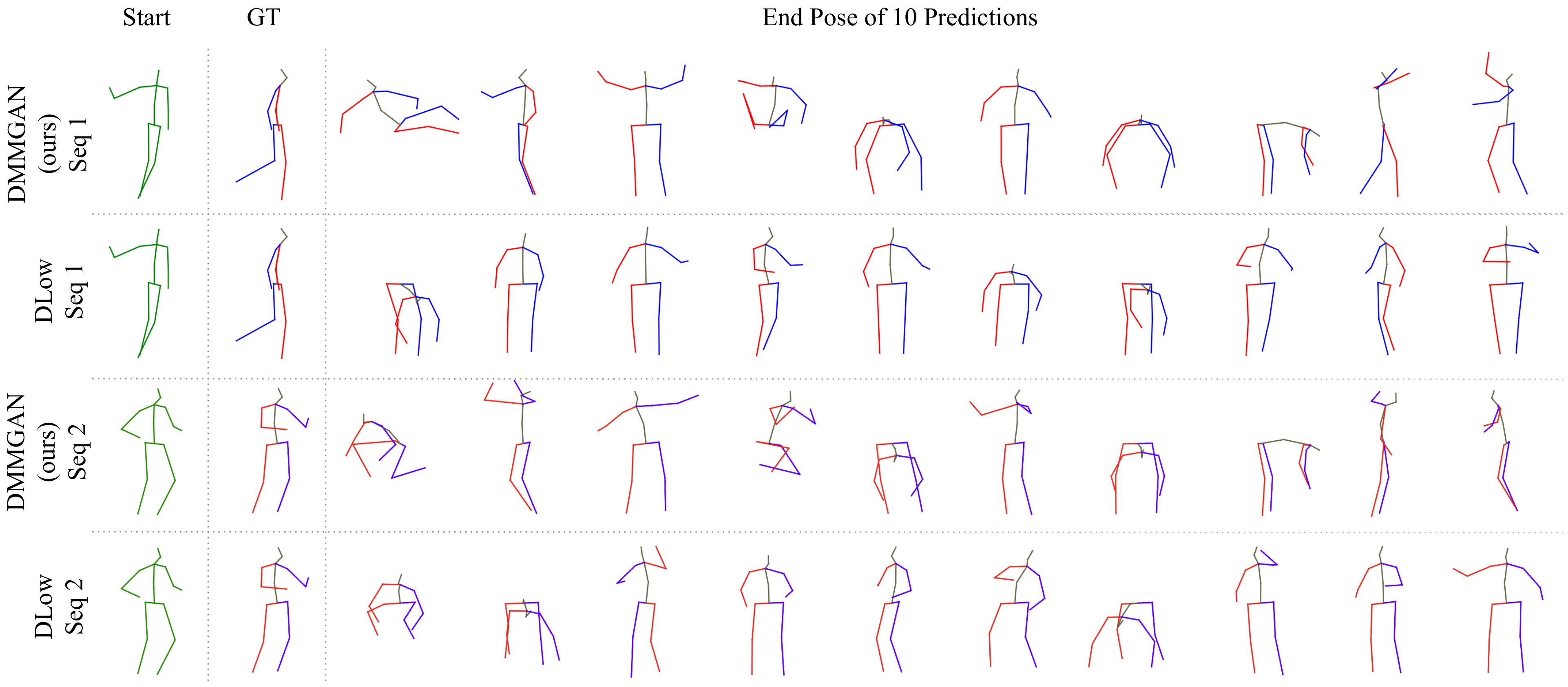}

\caption{Qualitative results of 3D pose predictions comparing our method, DMMGAN, to DLow in terms of diversity.}
\label{fig:pose3d}
\end{figure*}
\begin{figure*}[ht!]
\centering

\includegraphics[width=0.99\linewidth]{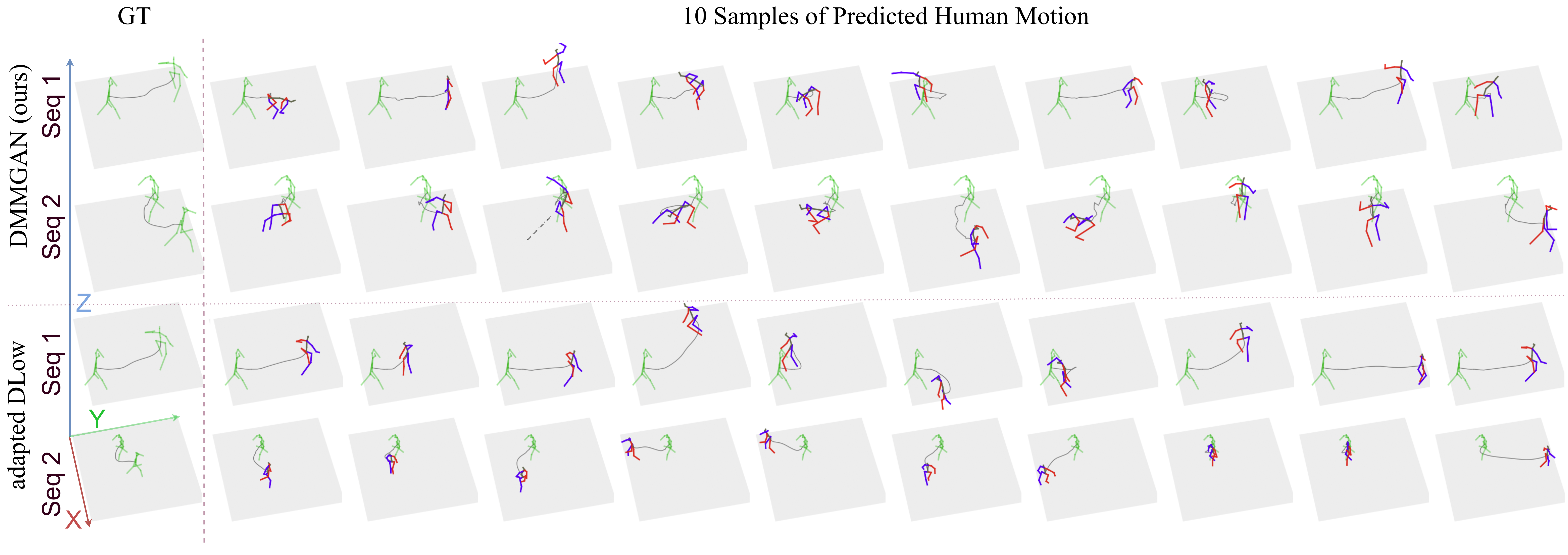}

\caption{Qualitative results of 3D motion predictions comparing our method, to DLow in terms of diversity.}
\label{fig:motion}
\end{figure*}
\begin{table*}[htbp]
  \centering
  \begin{tabular}{cccccc}
    Approach & APD  $\uparrow$  & ADE (m) $\downarrow$  & FDE (m) $\downarrow$ & MADE (m) $\downarrow$ & MFDE (m) $\downarrow$ \\
     &     & \begin{tabular}{c c}Pose &   Trajectory\end{tabular}   & \begin{tabular}{c c}Pose &   Trajectory\end{tabular}  & \begin{tabular}{c c}Pose &   Trajectory\end{tabular}   & \begin{tabular}{c c}Pose &   Trajectory\end{tabular}   \\
    \hline
    
    Adapted DLow & $5.55$ & 
    \begin{tabular}{c c} $0.483$ &   $0.195$\end{tabular} & 
    \begin{tabular}{c c} $0.621$ &   $0.457$\end{tabular} &
    \begin{tabular}{c c} $0.563$ &   $0.306$\end{tabular} &
    \begin{tabular}{c c} $0.649$ &   $0.553$\end{tabular} \\
    
    STPOTR & NA & 
    \begin{tabular}{c c} $0.507$ &   $0.139$\end{tabular} & 
    \begin{tabular}{c c} $0.758$ &  $0.277$\end{tabular} &
    \begin{tabular}{c c} NA &   NA\end{tabular} &
    \begin{tabular}{c c} NA &   NA\end{tabular} \\
    \hline
    ours:\\
    DMMGAN (ours) & \bm{$5.81$} & 
    \begin{tabular}{c c}$0.443$ &   $0.122$\end{tabular} & 
    \begin{tabular}{c c}$0.520$ &  $0.228$\end{tabular} &
    \begin{tabular}{c c}\bm{$0.540$} &   \bm{$0.192$}\end{tabular} &
    \begin{tabular}{c c}\bm{$0.597$} &   \bm{$0.342$}\end{tabular} \\
    MMGAN & $2.01$& 
    \begin{tabular}{c c}\bm{$0.422$} &   \bm{$0.104$}\end{tabular} & 
    \begin{tabular}{c c}\bm{$0.494$} &   \bm{$0.190$}\end{tabular} &
    \begin{tabular}{c c}$0.589$ &   $0.198$\end{tabular} &
    \begin{tabular}{c c}$0.665$ &   $0.360$\end{tabular} \\
    HipOnly & NA & 
    \begin{tabular}{c c} NA &   $0.156$\end{tabular} & 
    \begin{tabular}{c c} NA &   $0.306$\end{tabular} &
    \begin{tabular}{c c}NA &   NA\end{tabular} &
    \begin{tabular}{c c}NA &   NA\end{tabular} \\
  \end{tabular}
  
  \caption{Comparison of our systems versus two baselines for the full 3D motion experiment.}\label{tab:straight_real}
      \vspace{-5pt}
\label{tab:exp2}
\end{table*}

\subsection{3D Pose Experiment}

In the first experiment, we evaluate our 3D Pose generation module. Here, we compare our method against two baselines. The first one is DLow \cite{yuan2020dlow}, the state-of-the-art in diverse human 3D pose forecasting which outperforms all the currently known methods to the best of our knowledge. The authors of DLow \cite{yuan2020dlow} provide detailed comparisons to several other methods, which we will omit in this paper for brevity. The second baseline is STPOTR \cite{stopr}, a more recent method that also focuses on the 3D human motion prediction for robotics applications. STPOTR predicts only one future motion so we cannot use it for multi-modal evaluation.

Table \ref{tab:exp1} shows the results of this experiment. Our method outperforms both of the baselines and achieves the highest diversity while keeping both ADE and FDE lowest. Our method also has the highest coverage of the multi-modal ground-truth (MADE and MFDE). Also, we visually evaluate our method against DLow, in Fig. \ref{fig:pose3d}, we visualize the 10 end poses of our predictions versus the DLow for 2 random samples.  In both methods, we can see a comparable accuracy against the ground-truth data (GT). Although the diversity of our method is close to DLow, closer examination of Seq 1 shows that our method predicted sitting down, crouching, lying down, walking left and right, while DLow has qualitatively less diverse samples. 

\subsection{Full 3D Motion Experiment}
The second experiment evaluates our full system. In order to compare our system with a state-of-the-art diverse 3D motion model, we repurposed and retrained DLow \cite{yuan2020dlow} to forecast the human's trajectory by adding the hip joint to the joints \textbf{Adapted DLow} predicts. We also compare our system with STPOTR \cite{stopr}, which is one of the few papers that provides full 3D motion (pose and hip) prediction. We also include two variations of our models as an ablation study. The first model is \textbf{MMGAN} which is our full system trained without the \textit{similarity loss} and the second one is called \textbf{HipOnly} which is our \textit{Hip Prediction} module without the 3D pose prediction inputs. The HipOnly model evaluates the impact of the predicted 3D pose data on the accuracy of the trajectory estimation. (Fig. \ref{fig:pelvis_module}b without the right 3D pose Transformer encoder).

Based on the result of this experiment (Table \ref{tab:exp2}), our method outperforms the baselines by achieving the highest diversity while keeping the ADE and FDE lowest. In Fig. \ref{fig:motion}, we compare our prediction versus Adapted DLow and the ground-truth (GT) qualitatively \footnote{Please refer to \href{https://youtu.be/osJuFbtJsMg}{https://youtu.be/osJuFbtJsMg} for more examples.}. In these examples, Adapted DLow predicted only walking movement while DMMGAN could capture more diverse motions.

The \textit{HipOnly} model achieved a higher FDE and ADE compared to our model, which shows the benefit of using attention-based 3D pose generator during the trajectory forecasting. The results also highlight the impact of the \textit{similarity loss} on diversity of the predicted 3D motions. Our model without the similarity loss, MMGAN, achieved APD of 2 versus 5.8 for our full system. It is interesting to note that by removing the \textit{similarity loss}, model achieves a lower ADE and FDE with the cost of less diverse predictions.